\pdfoutput=1

\documentclass[11pt]{article}

\usepackage[final]{acl}

\usepackage{times}
\usepackage{latexsym}
\usepackage{hyperref}
\usepackage{amsmath}
\usepackage{multirow}
\usepackage{cleveref}
\usepackage{graphicx}
\usepackage{etoolbox}
\usepackage{colortbl}
\usepackage{spverbatim}
\usepackage{booktabs} 

\usepackage{times}
\usepackage{latexsym}
\usepackage{amsmath}
\usepackage{makecell}
\usepackage[T1]{fontenc}
\usepackage{hyperref}

\usepackage[utf8]{inputenc}

\usepackage{microtype}

\usepackage{inconsolata}
\usepackage{soul}

\usepackage{graphicx}

\newcommand{\ie}{\textit{i}.\textit{e}., }

\newcommand{\eg}{\textit{e.g., }}

\newcommand\icon{\raisebox{-10pt}{\includegraphics[width=2em]{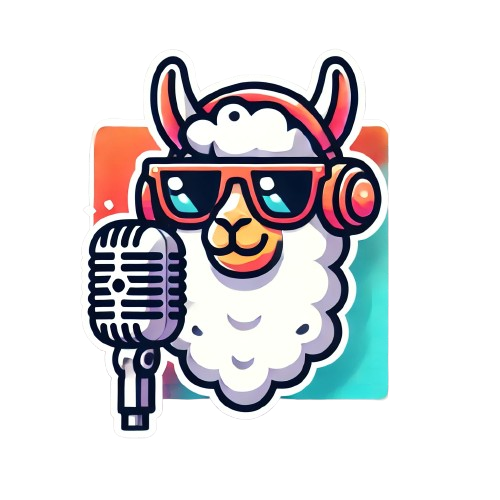}}}

\title{\icon OmniCharacter:
Towards Immersive Role-Playing Agents with Seamless Speech-Language Personality Interaction 
}

\author{
\textbf{Haonan Zhang}\textsuperscript{1,2}\thanks{Equal contribution.}\thanks{Work was done when interned at Tongyi Laboratory.},
\textbf{Run Luo}\textsuperscript{2,3,5}$^*$,
\textbf{Xiong Liu}\textsuperscript{2}$^*$, 
\textbf{Yuchuan Wu}\textsuperscript{2}, 
\textbf{Ting-En Lin}\textsuperscript{2}, 
\textbf{Pengpeng Zeng}\textsuperscript{1}, \\
\textbf{Qiang Qu}\textsuperscript{3,5}, 
\textbf{Feiteng Fang}\textsuperscript{2,5}, 
\textbf{Min Yang}\textsuperscript{3,5}, 
\textbf{Lianli Gao}\textsuperscript{4}, 
\textbf{Jingkuan Song}\textsuperscript{1}$^\ddag$,
\textbf{Fei Huang}\textsuperscript{2},
\textbf{Yongbin Li}\textsuperscript{2}\thanks{Jingkuan Song and Yongbin Li are corresponding authors.} 
\\
\textsuperscript{1}Tongji University \quad
\textsuperscript{2}Tongyi Laboratory \quad 
\textsuperscript{3}University of Chinese Academy of Sciences \\
\textsuperscript{4}Independent Researcher \quad
\textsuperscript{5}Shenzhen Institute of Advanced Technology, Chinese Academy of Sciences \\
\texttt{\{zchiowal, jingkuan.song\}@gmail.com} \quad \texttt{shuide.lyb@alibaba-inc.com} \\
}
\begin{document}
\maketitle

\begin{abstract}

Role-Playing Agents (RPAs), benefiting from large language models, is an emerging interactive AI system that simulates roles or characters with diverse personalities. However, existing methods primarily focus on mimicking dialogues among roles in textual form, neglecting the role's voice traits (\eg voice style and emotions) as playing a crucial effect in interaction, which tends to be more immersive experiences in realistic scenarios. Towards this goal, we propose \textit{\textbf{OmniCharacter}}, a first seamless speech-language personality interaction model to achieve immersive RPAs with low latency. Specifically, OmniCharacter enables agents to consistently exhibit role-specific personality traits and vocal traits throughout the interaction, enabling a mixture of speech and language responses. To align the model with speech-language scenarios, we construct a dataset named \textbf{\textit{OmniCharacter-10K}}, which involves more distinctive characters (20), richly contextualized multi-round dialogue (10K), and dynamic speech response (135K). 
Experimental results showcase that our method yields better responses in terms of both content and style compared to existing RPAs and mainstream speech-language models, with a response latency as low as 289ms. \footnote{Code and dataset are available at \url{https://github.com/AlibabaResearch/DAMO-ConvAI/tree/main/OmniCharacter}.}

\end{abstract}
\begin{figure}[]
  \centering
\includegraphics[width=\columnwidth]{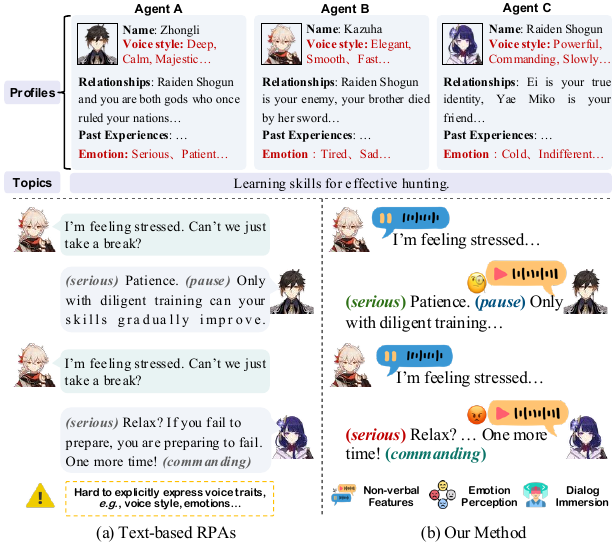}
  \caption{(a) Existing role-playing agents (RPAs) concentrate on engaging dialogue in textual form, whereas the role's voice traits are often ignored. (b) Our method considers the importance of the characters' vocal persona, \eg voice style and emotions, delivering a more seamless and immersive speech-language interaction.}
  \label{intro}
\end{figure}
\section{Introduction}

The rapid advancement of Large language models (LLMs)~\citep{wang2024survey,park2023generative,openai2022openai,li2022expansion} has demonstrated strong potential in interactive AI, enabling more natural and engaging human-computer interactions~\cite{luo2024mmevol,zhang2023spt}. One of the most promising and widely adopted research directions is the development of role-playing agents (RPAs)~\citep{wang2024characterbox,tu2024charactereval,shao2023character}, 
which have been applied in various domains, such as virtual assistants, AI-driven storytelling, and intelligent non-player characters (NPCs) in video games. 
The focus of RPAs allows LLMs to simulate diverse personas and mimic human-like behavior via predefined role profiles. 

Despite significant advances in this field, most existing studies primarily concentrate on replicating dialogues solely in textual form, while often overlooking the role’s crucial voice traits such as tone, style, emotions, and pauses.
The voice traits are important paralinguistic concepts that deeply reveal individual differences in conversations, enabling a more immersive and personalized role-playing interaction. 
As shown in~\Cref{intro} (a), for the same user query, although both agents provide reasonable responses, their unique tones, \ie Agent A’s serious yet calmed and Agent B’s angry and commanding, are difficult to convey through text alone, limiting their role-specific expression. Therefore, incorporating personalized voice traits into the current RPAs can help create more realistic and immersive interactions.

To bridge this gap, we introduce \textbf{\textit{OmniCharacter}}, a first step in realizing speech-language collaborative RPAs for immersive and seamless interaction. Specifically, a \textbf{Speech-Language Collaborative Model} is first built as a base model to perceive both language inputs (\ie role profile, dialogue contexts, and user text input) and speech input (\ie user speech input) for semantic alignment. 
Based on this, we propose a \textbf{Role Speech Decoder} to generate role-specific speech responses containing their unique voice traits. In particular, it mainly consists of two major components: 1) 
Role-context Guided Speech Token Prediction, which leverages the textual representations from LLM as prior, ensuring the alignment of discrete speech token generation with the corresponding textual contexts. 2) Role-aware Speech Synthesis, which generates waveform containing rich role-related voice traits from speech tokens specific to the given conditions (\ie profile, dialogue contexts, current query, and speaker embedding). 
Note that the proposed model can auto-regressively generate text and speech responses with 289ms latency, ensuring a seamless dialogue interaction.

To facilitate the development of OmniCharacter, we further present the \textbf{\textit{OmniCharacter-10K}}, a speech-language RPAs dataset with detailed profile annotations, diverse dialogues, and vivid audio responses tailored to each role's characteristic.
This dataset presents several appealing properties: 
1) \textbf{\textit{Large Vocabulary}}: 
It includes a total of 20 characters along with 10, 072 multi-turn dialogues, along with 135K audio responses.
2) \textbf{\textit{Rich Annotations}}: 
Each character is accompanied by a rich profile, highlighting aspects such as personality, voice style, relationships, and past experiences. Further, each dialogue turn is annotated with corresponding speech responses for both users and characters via TTS models~\citep{du2024cosyvoice,kim2021conditional}.
3) \textbf{\textit{Dynamic Curation}}:
The dialogue data is generated through interactive conversations between two chatbots, guided by the predefined profiles.
The main contributions are as follows:
\begin{itemize}
    \item We introduce \textbf{\textit{OmniCharacter}}, a pioneering step in realizing speech-language collaborative RPAs. 

    \item We construct \textbf{\textit{OmniCharacter-10K}}, a multi-turn dialogue dataset with detailed role profiles, conversion data, and high-quality audio annotations for advancing RPAs development.
    
    \item Experimental results demonstrate that our method achieves superior performance on the role-playing benchmark, \ie CharacterEval and SocialBench as well as providing robust results on general speech benchmarks, \ie LibriSpeech, AISHELL-2, and LibriTTS.
\end{itemize}

\begin{figure*}[]
  \centering
\includegraphics[width=\textwidth]{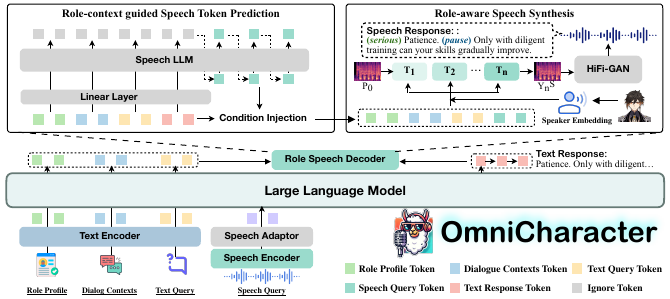}
  \caption{\textbf{The overview of our OmniCharacter framework}. We first build a speech-language collaborative model, a large language model that receives both speech and language inputs for unified modeling. Furthermore, we propose a role speech decoder to synthesize speech responses containing vocal traits of different characters by devising two innovative modules: \romannumeral1) Role-context Guided Speech Token Prediction, which aims to enhance the generation of speech tokens by leveraging the textual representations of the base model as a contextual prior. \romannumeral2) Role-aware Speech Synthesis, which generates speech responses from speech tokens maintaining unique character personality and voice traits based on several conditions.}
  \label{framework}
\end{figure*}

\section{Methodology}
In this section, we present an in-depth presentation of OmniCharacter, the first RPAs that perceive both speech and language understanding while generating character-specific audio-text responses with low latency. The framework is illustrated in~\Cref{framework}.

\subsection{Overview}
The traditional RPAs are designed to make LLMs interact with users or other agents by emulating specific characters.
To achieve this, RPAs utilize a profile denoted as $P$, along with the ongoing dialogue contexts $C_n = [q_1, r_1, ..., q_n]$ to generate a response $r_n$ consistently with character sets:
\begin{equation}
    \begin{aligned}
    r_n = \texttt{RPAs}(C_n, P),
    \end{aligned}
    \label{eq1}
\end{equation}
where $q_i$ represent the $i$-th input query and $r_i$ indicates the corresponding response. 
However, this paradigm relies on LLMs' behavior cloning ability and focuses mainly on textual form, neglecting essential vocal traits like voice style, tone, and emotions of different characters. 

To bridge this gap, we introduce OmniCharacter, the first model considering the role-related vocal persona by collaborating speech and language for immersive interaction. Specifically, OmniCharacter mainly contains two components: (1) a Speech-Language Collaborative model to align semantics by processing both language inputs (role profile, dialogue contexts, and text query) and speech input, (2) a Role Speech Decoder that aims to generate role-specific speech responses that reflect their unique voice traits. 
Thereby, the dialogues contexts $C_n$ can be redefined as $U_n = [X_1, Y_1, \dots, X_n]$, where $X_i \in \{X_i^S, X_i^T, X_i^{S+T}\}$ represents the $i$-th input query, supporting three types: audio-only ($X_i^S$), text-only ($X_i^T$), or audio-text ($X_i^{S+T}$), resulting in flexible interaction during dialogue.
Similarly, $Y_n\in \{Y_n^S, Y_n^T\}$ indicates the response, owning of two types, \ie audio response ($Y_n^S$) and text response ($Y_n^T$), which maintains the consistent and distinct personality of characters. Next, we elaborate on the details of two components in the subsequent subsections.

\subsection{Speech-Language Collaborative Model}
To fully align speech and language semantics, we first devise a unified speech-language collaborative model. Specifically, it comprises a speech encoder, a speech adaptor, and an LLM.

\noindent\textbf{Speech Encoder.} 
To tackle user or character speech input, we adopt a speech encoder to encode the speech input $X_n^S$ of the $n$-th dialogue turn to speech sequence $M_n^S = [M_{n, 1}^S, ..., M_{n, N_f}^S]$, where $N_f$ indicate the number of audio frames. 

\noindent\textbf{Speech Adaptor.}
To address the high temporal redundancy of the encoded speech sequence above, a down-sampling strategy is further applied by grouping every $k$ consecutive frame into compact sequence $Z_n^S = [Z_{n,1}^S,...,Z_{n,\left\lfloor N_s/k \right\rfloor}^S]$, where
\begin{equation}
    \begin{aligned}
    Z_{n,i}^S = M_{n, {k*i}}^S \oplus M_{n, {k*i+1}}^S \oplus...\oplus M_{n, {k*i+k-1}}^S.
    \end{aligned}
\end{equation}
This sequence is then encoded by a speech adapter $\tau$ to make it compatible with the LLM embedding:
\begin{equation}
    \begin{aligned}
    E_n^S = \tau(Z_n^S).
    \end{aligned}
\end{equation}

\noindent\textbf{Large Language Model.}
We utilize Qwen2.5-7B-Instruct~\citep{yang2024qwen2} as the LLM, a state-of-the-art large language model renowned for its advanced reasoning capabilities to align text and audio sequences for unified modeling. In particular, we process the text inputs, including the role profile $P$, dialogue contexts $U_n$, and the text query $X_n^T$ to form an overall sequence $E_n^T=\{E_{n,i}^T\}_{i=1}^{N_t}$ via a text encoder. This sequence is then concatenated with the speech sequence $E_n^S=\{E_{n,i}^S\}_{i=1}^{N_s}$ and sent into LLM. The $N_t$ and $N_s$ indicate the length of text and speech sequences respectively. The prompt template for organizing the overall inputs is showcased in the~\Cref{prompt}. Finally, the LLM is trained via an auto-regressive manner:
\begin{equation}
    \begin{aligned}
    \mathcal{L}_{language} = -\sum_{i=1}^{N_t} \log P(O^T_i \mid O^T_{<i}),
    \end{aligned}
    \label{language}
\end{equation}
where $O^T_i$ represents the textual token output from the LLM.

\subsection{Role Speech Decoder}

\noindent\textbf{Role-context Guided Speech Token Prediction.}
Intuitively, it is straightforward to use the LLM to predict both text tokens $O^T$ and speech tokens $O^S$. However, we find this strategy makes the model difficult to train, causing duplicate speech token outputs and hallucinations that are inconsistent with the semantics of text tokens.
To facilitate stable and accurate speech token prediction, we propose an innovative module known as role-context guided speech token prediction built upon the speech-language collaborative model.
Specifically, it employs a lightweight language model (denoted as SpeechLLM) that effectively utilizes the context representations $H = [H_1, ..., H_{N_s+N_t}]$ output from LLM to generate speech tokens:
\begin{equation}
    \begin{aligned}
    O^S &= \text{SpeechLLM}(O_m^S | O^S_{1:m-1}, \phi(H)),
    \end{aligned}
\end{equation}
where $\phi$ is a linear projection layer to map the context representations into the SpeechLLM space. We find that this design not only stabilized the training but also ensured that the generated speech tokens are highly aligned with the contextual semantics, effectively preventing the speech token repetitive output problem. 
Note that the SpeechLLM also follows an auto-regressive training objective:
\begin{equation}
    \begin{aligned}
    \mathcal{L}_{speech} = -\sum_{i=1}^{N_s} \log P(O^S_i \mid O^S_{<i}).
    \end{aligned}
    \label{speech}
\end{equation}

\begin{figure*}[]
  \centering
\includegraphics[width=\textwidth]{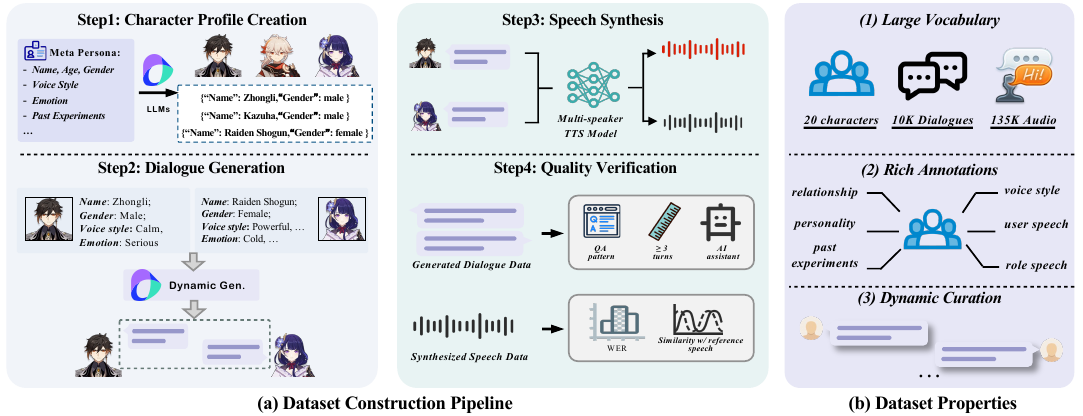}
  \caption{\textbf{Illustration of OmniCharacter-10K.} (a) Dataset Construction Pipeline, which consists of four steps: (1) Characters Profile Creation, (2) Dialogue Generation, (3) Speech Synthesis, and (4) Quality Verification. (b) Dataset Properties, which has three appealing properties of our dataset: (1) Large Vocabulary, the dataset includes 20 characters, 10K multi-turn dialogues, and 135K audio responses, (2) Rich Annotations, each character is equipped with rich text and speech annotations, and (3) Dynamic Curation, we generate dialogue data through chatbot interactions based on predefined character profiles.}
  \label{data}
\end{figure*}

\noindent\textbf{Role-aware Speech Synthesis.}
A high-quality speech response that reflects the character's voice traits and persona is important for immersive RPAs. To achieve this, we propose a role-aware speech synthesis module, which generates the audio response containing voice traits of characters from the speech tokens. Following~\citep{zeng2024glm}, we first decode the speech tokens into the Mel spectrogram and then synthesize the waveform with the generated Mel spectrogram as input.
To achieve this, we adopt a conditional flow matching (CFM) model to sample the Mel spectrogram specified by the context representations $H$, speech tokens $O^S$, and speaker embedding $v$~\citep{cam++} extracted from character voice as conditions. The sampling process can be defined by a time-dependent vector field $T=\{T_1,...,T_n\}$.
To improve efficiency, the optimal-transport (OT) flow is used. We summarize the above process as:
\begin{equation}
    \begin{aligned}
    Y_n^S &= \text{OT-CFM}(p_0 | H, v, O^S), \\
    \end{aligned}
\end{equation}
where $p_0$ is the initial Mel spectrogram. Ultimately, after obtaining the generated Mel spectrogram, a HiFi-GAN vocoder serves to synthesize a continuous waveform from the Mel spectrogram, facilitating the conversion of spectral representations into high-fidelity audio signals $Y^S_n$ containing characters' voice traits such as voice style and emotions.

\subsection{Training Strategy}  \label{train}
Following~\citep{fang2024llama}, we adopt a two-stage training strategy for OmniCharacter. In the first stage, we train the speech adapter and LLM to generate text responses based on the text and speech inputs using the objective $\mathcal{L}_{language}$ in~\Cref{language}. In the second stage, we only train the linear layer and speech LLM in role-context guided speech token prediction module for speech token prediction by
using the objective $\mathcal{L}_{speech}$ in~\Cref{speech}. For role-aware speech synthesis, we utilize the pre-trained conditional flow matching and HiFi-GAN weights from GLM-4-Voice~\citep{zeng2024glm} and fine-tune them on high-quality role speech data\footnote{https://genshin.hoyoverse.com/en/}.

\section{The OmniCharacter-10K Dataset}
To facilitate the model with speech-language scenarios, we construct OmniCharacter-10K, a multi-turn dialogue dataset with distinctive characters with rich and expressive text and speech annotations.
In this section, we describe the data collection, annotation, and verification of OmniCharacter-10K. We also introduce the statistics and distribution of it. The illustration of OmniCharacter-10K is showcased in~\Cref{data}. 

\subsection{Data Collection, Annotation, and Verification}

\noindent\textbf{Step-1: Character Profile Creation.} To make sure the collected characters have distinct traits as well as contain high-quality audio, we select from games since (1) the availability of rich information (\eg personality, background, and vocal tone), and (2) clear, noise-free character audio recorded by professional voice actors. We end up with 20 foundational characters (10 Chinese and 10 English) from Genshin Impact to construct our dataset. 
Subsequently, we use a large language model to summarize the meta information for each character, and then instruct this model to expand the information to ensure the diversity and complexity. At last, all character profiles are subjected to rigorous human checks to ensure accuracy and reliability. Moreover, given the diversity of humans and the open-domain nature of the question, we do not consider the profile for the `user' side.

\noindent\textbf{Step-2: Dialogue Generation.} 
After selecting the characters and defining their profiles, we generate multi-turn dialogues to evaluate role-playing consistency and conversational quality. Specifically, we deploy two large language models in a simulated interaction: one assumes the role of the target character, while the other acts as a user or a different character. Each model is conditioned on its respective profile, which includes key personality traits, background information, emotions, etc. The dialogue is guided by prompts that encourage natural and persona-consistent responses across multiple turns. This setup enables us to systematically assess how well the models embody their assigned roles, maintain character coherence, and engage in meaningful, contextually appropriate exchanges.

\begin{table}[]
\resizebox{\linewidth}{!}{%
\begin{tabular}{lccccc}
\toprule
\textbf{Splits} & \# \textbf{Characters}  & \textbf{Avg. Turns/Conv.} & \# \textbf{Samples} & \# \textbf{Speech Hours.} (\textbf{user/character}) \\ \midrule
Training & 20  & 15 & 9, 672 & 360.3 (176.4/183.9) \\
Test & 20  & 15  & 400 & 14.84 (7.15/7.69) \\ \bottomrule
\end{tabular}%
}
\caption{The statistic of OmniCharacter-10K dataset.}
\label{stat}
\end{table}

\begin{figure}[]
  \centering
\includegraphics[width=\linewidth]{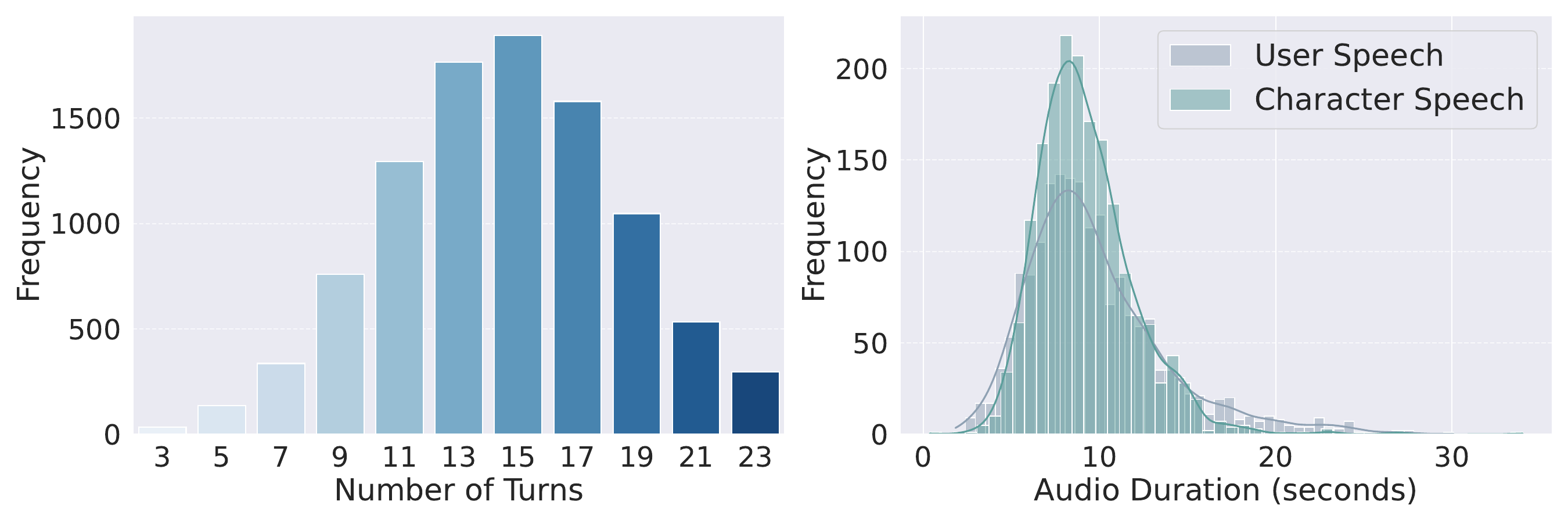}
  \caption{\textbf{Left:} Distribution of dialogue turns across samples in the OmniCharacter-10K dataset. \textbf{Right:} Distribution of audio duration for user speech and character speech respectively.}
  \label{dis}
\end{figure}

\noindent\textbf{Step-3: Speech Synthesis.}
The generated multi-turn dialogues consist of open-domain text without corresponding audio annotations. To handle this, we first collected 40K high-quality audio samples for 20 characters, which are professionally recorded by voice actors and extracted from the game’s asset files. Then, we utilize these audios as seed data and train a multi-speaker text-to-speech (TTS) model, \ie VITS~\citep{kim2021conditional} for the audio synthesis of open-text in dialogue. For the user side, we use the advanced TTS model CosyVoice~\citep{du2024cosyvoice} to synthesize audio. Each dialogue sample has a 50\% ratio of being assigned a generic male or female voice, ensuring its balance and diverse distribution.

\begin{table*}[]
\centering
\resizebox{\textwidth}{!}{%
\begin{tabular}{lcccccccccccccccc}
\toprule
\multirow{2}{*}{\textbf{Models}} & \multicolumn{6}{c}{\textbf{Character Consistency}} & \multicolumn{4}{c}{\textbf{Conversational Ability}} & \multicolumn{5}{c}{\textbf{Role-playing Attractiveness}} & \multirow{2}{*}{\textbf{Avg.}} \\ \cmidrule(lr){2-7} \cmidrule(lr){8-11} \cmidrule(lr){12-16} 
 & KE & KA & KH & PB & PU & \textbf{Avg.} & Flu. & Coh. & Cons. & \textbf{Avg.} & HL & CS & ED & Emp. & \textbf{Avg.} & \\ \midrule

\textbf{\textit{Proprietary Models}} & & & & & & & & & & & & & & & &  \\
BC-NPC-Turbo & 1.802  & 2.964 & 2.993 & 2.910 & 3.151 & 2.764 & 3.578 & 3.898 & 3.916 & 3.798 & 3.836 & 2.643 & 2.336 & 2.971 & 2.946  & 3.169 \\
MiniMax & 1.835   & 2.910   & 2.944   & 2.774   & 3.125            & 2.718  & 3.609           & 3.932               & 3.811            & 3.784          & 3.768               & 2.672      & 2.150      & 3.017      & 2.902 & 3.134 \\ 
GPT-3.5 & 1.716   & 2.339   & 2.212   & 1.921   & 2.316   & 2.101   & 2.629      & 2.917      & 2.700            & 2.749          & 2.565      & 2.422      & 1.660      & 2.526      & 2.293  & 2.381 \\ 
GPT-4 & 2.250         & 2.855   & 2.785   & 2.721   & 2.873   & 2.697  & 3.332      & 3.669      & 3.343            & 3.448          & 3.143      & 3.184            & 2.153      & 3.010      & 2.873 & 3.006 \\
\midrule
\textbf{\textit{Open-sourced Models}} & & & & & & & & & & & & & & & & \\
ChatGLM3-6B & 2.016 & 2.792 & 2.704 & 2.455 & 2.812 & 2.556 & 3.269 & 3.647 & 3.283 & 3.399 & 3.064 & 2.932 & 1.969 & 2.993 & 2.739 & 2.898 \\
Baichuan2-7B  & 1.813  & 2.849  & 2.929  & 2.830 & 3.081 & 2.700  & 3.551 & 3.894 & 3.827 & 3.757 & 3.670 & 2.728 & 2.115 & 2.984  & 2.874 & 3.110 \\
Baichuan2-13B & 1.802 & 2.869 & 2.946    & 2.808 & 3.081 & 2.701   & 3.596       & 3.924    & 3.864  & 3.759    & 3.700    & 2.703    & 2.136    & 3.021       & 2.890  & 3.116 \\
InternLM-7B   & 1.782 & 2.800 & 2.781 & 2.719 & 3.016 & 2.620  & 3.527    & 3.823    & 3.744  & 3.698  & 3.546    & 2.622    & 2.070    & 2.897    & 2.784  & 2.983 \\
InternLM-20B  & 1.945 & 2.916    & 2.920 & 2.753 & 3.041 & 2.715  & 3.576    & 3.943    & 3.717  & 3.745         & 3.582    & 2.885    & 2.132    & 3.047    & 2.911  & 3.123 \\
CharacterGLM  & 1.640 & 2.819 & 2.738 & 2.301 & 2.969 & 2.493  & 3.414    & 3.717    & 3.737  & 3.623  & 3.738    & 2.265    & 1.966    & 2.812    & 2.695 & 2.937 \\
Llama-3.1-8B & 2.197 & 2.701 & 2.615 & 3.130 & 2.704 & 2.669 & 3.059 & 3.477 & 3.071 & 3.202 & 2.922 & 2.934 & 2.634 & 2.759 & 2.812 & 2.894  \\ 

Qwen-7B & 1.956 & 2.728 & 2.633 & 2.605 & 2.780 & 2.540  & 3.187 & 3.564 & 3.229       & 3.327  & 3.036 & 2.791 & 2.052 & 2.838 & 2.679  & 2.848 \\ 
Qwen-14B & 1.988 & 2.800 & 2.811 & 2.744 & 2.900 & 2.649  & 3.351    & 3.765    & 3.510  & 3.542  & 3.354    & 2.871    & 2.237       & 2.970    & 2.858 & 3.016 \\ 
Qwen2-7B-Instruct & 1.966  & 2.537  & 2.412  & 2.313  & 2.436  & 2.333  & 2.864  & 3.171 & 2.743  & 2.926  & 2.655 & 2.612 & 1.867 & 2.654 & 2.447  &  2.569 \\ 

Qwen2.5-7B-Instruct & 2.207  & 2.740  & 2.633  & 2.700  & 2.614  & 2.579   & 3.125  & 3.401 & 2.971  & 3.166   & 2.786 & 2.871 & 2.180 & 2.826 & 2.666  & 2.804  \\ 

Qwen2.5-7B-Instruct\dag & 2.172  & 3.012  & 2.887  & 3.526  & 2.879  & 2.895   & 3.321  & 3.657 & 3.326  & 3.434   & 3.278  & 3.127  &  2.728 & 3.217 & 3.087 & 3.138  \\ 

\midrule

\cellcolor{gray!15}\textbf{OmniCharacter \textit{w/o} audio} & \cellcolor{gray!15}2.188  & \cellcolor{gray!15}3.024  & \cellcolor{gray!15}2.926  & \cellcolor{gray!15}3.547  & \cellcolor{gray!15}2.955  & 
\cellcolor{gray!15}2.928     & \cellcolor{gray!15}3.402  & \cellcolor{gray!15}3.732 & \cellcolor{gray!15}3.432  & 
\cellcolor{gray!15}3.522 & \cellcolor{gray!15}3.303 & \cellcolor{gray!15}3.266 & \cellcolor{gray!15}2.933 & \cellcolor{gray!15}3.172 & \cellcolor{gray!15}3.168 & \cellcolor{gray!15}3.206 \\

\cellcolor{gray!15}\textbf{OmniCharacter} & \cellcolor{gray!15}2.230  & \cellcolor{gray!15}3.040  & \cellcolor{gray!15}2.918  & \cellcolor{gray!15}3.531  & \cellcolor{gray!15}2.988  & 
\cellcolor{gray!15}2.941      & \cellcolor{gray!15}3.369  & \cellcolor{gray!15}3.768 & \cellcolor{gray!15}3.410  & 
\cellcolor{gray!15}3.516 & \cellcolor{gray!15}3.374 & \cellcolor{gray!15}3.261 & \cellcolor{gray!15}3.002 & \cellcolor{gray!15}3.187 & \cellcolor{gray!15}3.206 & \cellcolor{gray!15}3.221 \\

\bottomrule
\end{tabular}%
}
\caption{\textbf{Performance comparison with state-of-the-art methods on \textit{CharacterEval}.} We evaluate the models across three dimensions including: 
(A) \textbf{Character Consistency}: (a-1) KE: Knowledge-Exposure, (a-2) KA: Knowledge-Accuracy, (a-3) KH: Knowledge-Hallucination, (a-4) PB: Persona-Behavior, (a-5) PU: Persona-Utterance. 
(B) \textbf{Conversational Ability}: (b-1) Flu.: Fluency, (b-2) Coh.: Coherency, (b-3) Cons.: Consistency. 
(C) \textbf{Role-playing Attractiveness}: (c-1) HL: Human-Likeness, (c-2) CS: Communication Skill, (c-3) ED: Expression Diversity, (c-4) Emp.: Empathy. \dag~means we finetune pretrained Qwen-2.5-7B-Instruct on the textual portion of OmniCharacter-10K.}
\label{tab1}
\end{table*}
\noindent\textbf{Step-4: Quality Verification.} To ensure the quality of the constructed dataset, we manually filter samples based on several criteria. For text data, we performed data filtering by retaining the samples following \texttt{ABAB} (user-role or role1-role2) pattern. Besides, we preserved dialogues longer than three turns, filtering out shorter ones. Furthermore, we remove data that mimics the style of general AI assistance such as `\textit{I am a helpful AI assistant...}', as well as unnecessary explanatory prefaces and postfaces. For speech data, we first use the Whisper-large-v3~\citep{radford2023robust} to convert the synthesized audio into text and compute the WER metrics with the corresponding text, where the WER greater than 10 are rejected. Then, we use WavLLM~\citep{hu2024wavllm} to compute the similarity between the synthesized audio and reference audio, where the similarity less than 0.8 threshold is also rejected.

\subsection{Statistics of OmniCharacter-10K}
\Cref{stat} presents the statistics of the OmniCharacter-10K dataset, which includes a total of 20 characters with 9, 672 training samples and 400 test samples.
The training and test sets are rich in speech annotations, with 360.3 hours and 14.84 hours of audio data, respectively. In addition, \Cref{dis} shows the distribution of dialogue turns (left) and audio duration (right) across the dataset. On one hand, more than 80\% of the dialogues are longer than 10 turns, highlighting the dataset's focus on multi-turn interactions. On the other hand, the audio duration statistics further reveal that the speech in the dataset is sufficiently long to enable immersive dialogue experiences. These statistics emphasize the dataset's capability to support the evaluation of RPAs in facilitating immersive dialogues. 

\begin{table*}[]
\resizebox{\textwidth}{!}{%
\begin{tabular}{lcccccccccc}
\toprule
\multirow{2}{*}{\textbf{Models}} & \multicolumn{6}{c}{\textbf{Individual Level}}                                                                       & \multicolumn{3}{c}{\textbf{Group Level}}         & \multirow{2}{*}{\textbf{Avg.}} \\ \cmidrule(lr){2-10}
                                 & SA Style & SA Know. & EP Situ. & EP Emo. & CM Short & CM Long & Pos.  & Neu.  & Neg.  &                               \\ \midrule
Qwen-72B-Chat                    & 83.87             & 90.64             & 53.10             & 52.89            & 83.29             & 73.15            & 91.53          & 73.44          & 63.82          & 73.97                         \\
LLaMA-2-70B-Chat                 & 67.61             & 70.78             & 35.74             & 38.47            & 45.57             & 26.74            & 69.87          & 45.29          & 39.37          & 48.83                         \\
Qwen-14B-Chat                    & 77.06             & 86.15             & 45.71             & 43.78            & 65.32             & 51.37            & 78.32          & 58.25          & 59.21          & 62.80                         \\
LLaMA-2-13B-Chat                 & 57.62             & 65.51             & 37.12             & 32.56            & 30.43             & 29.82            & 66.38          & 42.25          & 26.27          & 43.11                         \\ \midrule
LLaMA-2-7B-Chat                  & 48.76             & 51.23             & 31.23             & 28.91            & 25.38             & 21.89            & 44.98          & 24.19          & 27.67          & 33.80                         \\
Mistral-7B                       & 50.12             & 61.17             & 36.48             & 31.72            & 31.78             & 25.42            & 65.67          & 46.34          & 28.96          & 41.96                         \\
Qwen-7B-Chat                     & 66.44             & 71.16    & 41.68             & 40.68   & 67.45    & 53.45            & 75.61 & 52.78          & 43.11          & 56.93                         \\
\cellcolor{gray!15}OmniCharacter                    & \cellcolor{gray!15}70.93    & \cellcolor{gray!15}65.68             & \cellcolor{gray!15}58.03    & \cellcolor{gray!15}38.24            & \cellcolor{gray!15}59.76             & \cellcolor{gray!15}57.86   & \cellcolor{gray!15}69.11          & \cellcolor{gray!15}63.26 & \cellcolor{gray!15}43.98 & \cellcolor{gray!15}58.53                \\ \bottomrule
\end{tabular}%
}
\caption{\textbf{Performance comparison with state-of-the-art methods on SocialBench.} We evaluate the models across several dimensions including: (1) \textbf{Self-Awareness}: role style (SA Style) and role knowledge (SA Know.) (2) \textbf{Emotional Perception}: situation understanding (EP Situ.) and emotion detection (EP Emo.) (3) \textbf{Conversation Memory}: short-term (CM Short) and long-term (CM Long) (4) \textbf{Social Preference}: positive (Pos.), neutral (Neu.), and negative (Neg.).}
\label{socialbench}
\end{table*}

\section{Experiments}

\subsection{Experimental Setups} 

\noindent\textbf{Model Configuration.} 
We employ the Whisper-large-v3 as the speech encoder and 
Qwen-2.5-7B-Instruct as the LLM in speech-language collaborative model. 
In role-context guided speech token prediction, we employ the lightweight Qwen2.5-0.5B-Instruct~\citep{qwen2.5} model as SpeechLLM to generate speech tokens.
To facilitate this process, a speech tokenizer derived from GLM-4-Voice~\citep{zeng2024glm} with 16K vocabulary size is utilized to extract discrete speech tokens. 
We resample all speech data to a frequency of 16Khz.
The speech-language collaborative model and role-context guided speech token prediction are initialized using pre-trained parameters from OpenOmni~\citep{luo2025openomni}, and we conduct fine-tuning on the OmniCharacter-10K train set. The weights of role-aware speech synthesis are initialized from pretrained checkpoint of GLM-4-Voice.

\noindent\textbf{Training Details.}
The OmniCharacter employs a two-stage training strategy as described in~\Cref{train}. In the first stage, we utilize the AdamW optimizer with a warmup ratio of 0.3, gradually increasing the learning rate until it reaches 5e-4. The second stage retains the parameters established in the first stage, and we only adjusted the final learning rate to 5e-5. 
We set the batch size as 32 for both two stages.
The entire training process is finished within 3 epochs on 8$\times$A100 GPUs.

\noindent\textbf{Datasets.}  
The evaluation leverages the following datasets: CharacterEval~\citep{tu2024charactereval} and SocialBench~\citep{chen2024socialbench} are used to testify models' basic language understanding. OmniCharacter-10K test set is employed for character-aware speech-language evaluation to test speech generation and alignment abilities. LibriSpeech~\citep{panayotov2015librispeech}, AISHELL-2~\citep{du2018aishell}, and LibriTTS~\citep{zen2019libritts} are general speech datasets for model's generalization test. 

\begin{table}[]
\centering
\resizebox{\linewidth}{!}{%
\begin{tabular}{lccccccc}
\toprule
\multicolumn{1}{l}{\multirow{2}{*}{\textbf{Models}}} & \multicolumn{2}{c}{\textbf{S2TIF}} & \multicolumn{2}{c}{\textbf{S2SIF}}   \\ \cmidrule(l){2-5} 
\multicolumn{1}{c}{} & Content ↑ & Style ↑ & Content ↑ & Style ↑ \\ \midrule
SpeechGPT &1.25 &1.28 &1.79 &1.85   \\
LLaMA-Omni & 2.46 & 2.71 & 2.26 &2.38  \\
\cellcolor{gray!15}OmniCharacter & \cellcolor{gray!15}3.89 & \cellcolor{gray!15}4.18 & \cellcolor{gray!15}2.64 & \cellcolor{gray!15}2.55   \\ \bottomrule
\end{tabular}%
}
\caption{\textbf{Performance comparison with state-of-the-art methods on OmniCharacter-10K test split.} ChatGPT scores for S2TIF, S2SIF tasks, and alignment scores between speech and text responses.}
\label{tab2}
\end{table}

\subsection{Performance Comparison}

\noindent\textbf{Basic Language Understanding.} 
To evaluate the effectiveness of our OmniCharacter in basic language understanding, we compare our method with RPAs methods on CharacterEval~\citep{tu2024charactereval} benchmark as shown in~\Cref{tab1}.
Note that the characters tested in CharacterEval differ from those in our training data. 
The results illustrate that our model outperforms the proprietary and open-source models on most metrics. Particularly, compared with models specifically designed for role-playing task including BC-NPC-Turbo and MiniMax, our method achieves considerable improvements. Moreover, OmniCharacter surpasses open-sourced models with larger scales such as InterLM-20B and Qwen-14B, achieving superior performance in both character consistency and role-playing attractiveness. 
Further, we evaluate our method on SocialBench~\cite{chen2024socialbench} as shown in~\Cref{socialbench}. As observed, our method improves performance across most metrics when compared to the same-size models. Although OmniCharacter is not as optimal as larger-scale models, it still delivers comparable performance, securing the second-best position in terms of the average score.
The results presented above provide compelling evidence that the integration of audio modality enhances basic language understanding.

\noindent\textbf{Character-aware Speech-Language Evaluation.} To thoroughly evaluate the speech-language collaboration ability of OmniCharacter for immersive and personalized role-playing interaction, we conduct evaluations from two perspectives: metrics-based and human-based evaluations. 

The metrics-based evaluation consists of three tasks~\citep{fang2024llama}: (1) speech-to-text instruction-following (S2TIF): whether the model's text response correctly answers the input speech query; (2) speech-to-speech instruction-following (S2SIF): whether the model's speech response correctly answers the input speech query. We ask GPT-4o to score the response in a range from 1 to 5 from content and style aspects. As shown in~~\Cref {tab2}, our model consistently enhances performance across all metrics, achieving superior results compared to LLaMA-Omni and SpeechGPT. These results confirm that OmniCharacter is capable of effectively handling user or characters' instructions, while simultaneously generating consistent text and speech responses.

In terms of human-based evaluation, five experts are assigned to examine the model's outputs across six voice-related dimensions as shown in~\Cref{tabnew}. Each expert is asked to score the responses on a scale from 1 to 10 for each dimension. 
From the table, we figure out that our method reaches the best performance across all dimensions, proving the superiority of our model which is capable of providing immersive interactions for users.

\begin{table}[]
\centering
\resizebox{\linewidth}{!}{%
\begin{tabular}{lccccccc}
\toprule
\textbf{Models} & \textbf{Flu.} & \textbf{Cons.} & \textbf{Emo.} & \textbf{Cla.} & \textbf{App.} & \textbf{Imm.} \\ \midrule
SpeechGPT & 6.12 & 3.86  & 3.75 & 6.89 & 4.23 & 3.64  \\
LLaMA-Omni & 6.88 & 4.27 & 3.44 & 6.69 & 4.78  & 4.68 \\
\cellcolor{gray!15}OmniCharacter & \cellcolor{gray!15}7.97 & \cellcolor{gray!15}6.84 & \cellcolor{gray!15}6.23 & \cellcolor{gray!15}7.88 & \cellcolor{gray!15}5.63 & \cellcolor{gray!15}8.52 \\ \bottomrule
\end{tabular}%
}
\caption{\textbf{Performance comparison with state-of-the-art methods on OmniCharacter-10K test split.}
Human score (averaged from 5 experts) across six voice-related dimensions: fluency (Flu.), consistency (Cons.), emotional expression (Emo.), clarity (Cla), appropriateness (App.), and immersion (Imm.).}
\label{tabnew}
\end{table}

\noindent\textbf{Generalization on General Speech Benchmarks.}
To better prove the robust ability of OmniCharacter, we evaluated the performance of our model on three widely used speech datasets, LibriSpeech, AIShell-2, and LibriTTS, focusing on both Automatic Speech Recognition (ASR) and Text-to-Speech (TTS) tasks. As presented in \Cref{tab3}, our model achieves comparable performance compared with previous models, demonstrating its generalization ability.  
These results further highlight the robustness of OmniCharacter in generating coherent and accurate outputs, showcasing its potential in real-world speech applications.

\subsection{Impact of Audio Modality} 
To assess the impact of the audio modality on role-playing agents (RPAs), we conduct an ablation study by removing audio input from OmniCharacter and evaluating the model on the CharacterEval benchmark, as shown in~\Cref{tab4}. We report average scores across all metrics for clarity. The results indicate that incorporating audio significantly improves character consistency and conversational ability, while achieving comparable performance in role-playing attractiveness. These findings highlight the importance of audio in enhancing language understanding and contributing to more natural, expressive role-playing interactions.

\subsection{Similarity of Synthesized Voice for Characters}
\begin{figure}[]
  \centering
\includegraphics[width=\linewidth]{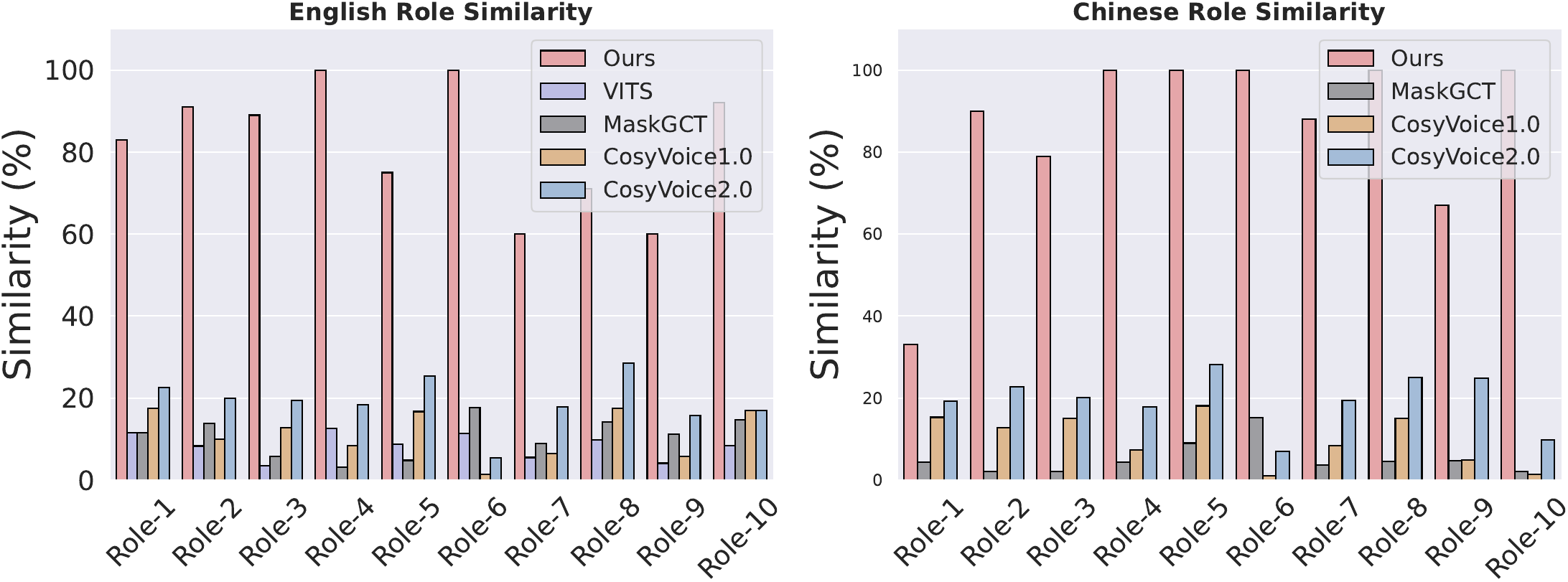}
\caption{\textbf{Comparison of character voice similarity on OmniCharacter-10K test set.} Our model generates audio responses that better match the timbre of characters.}
  \label{sim}
\end{figure}

To verify the model's capability of synthesizing character-specific speech containing their voice traits, we use WavLLM to compute the cosine similarity between the reference speech and the generated speech on OmniCharacter-10K test set, as shown in~\Cref{sim}. The comparison methods include state-of-the-art TTS systems, \ie VITS, MaskGCT, CosyVoice1.0, and CosyVoice2.0. For our model, we directly input the corresponding text to generate the speech output. For other models, we leverage their voice cloning capabilities by providing reference speech, along with the given text, to synthesize the corresponding speech. If the cosine similarity between the output speech and the reference speech is greater than 0.9, we classify them as belonging to the same character. As observed, our model consistently achieves higher cosine similarity for both Chinese and English characters' voices. These results highlight the effectiveness of our approach in preserving the voice traits of characters, achieving a more interactive experiment. 

\begin{table}[]
\centering
\resizebox{\linewidth}{!}{%
\begin{tabular}{lccccc}
\toprule
\multirow{2}{*}{\textbf{Models}} & \multicolumn{2}{c}{\textbf{LibriSpeech} (ASR-WER)} & \textbf{AISHELL-2} (ASR-CER) & \textbf{LibriTTS} (TTS-WER)  \\
 & test-clean & test-other & test & test-clean \\ \midrule
CosyVoice 1.0 & - & - & - & 3.17 \\
Whisper-large-v3 & 2.50 & 4.53 & - & - \\
Qwen2-Audio & 2.00 & 4.50  & 3.30 & -  \\
\cellcolor{gray!15}OmniCharacter &\cellcolor{gray!15}3.26 &\cellcolor{gray!15}4.23  &\cellcolor{gray!15}6.62  &\cellcolor{gray!15} 7.23 \\ \bottomrule
\end{tabular}%
}
\caption{\textbf{Performance comparison with the state-of-art methods on general speech benchmarks} including LibriSpeech, AISHELL-2, and LibriTTS. The model is evaluated on two tasks, \ie Automatic Speech Recognition (ASR) and Text-to-Speech (TTS). WER: Word Error Rate. CER: Character Error Rate. 
}
\label{tab3}
\end{table}

\begin{table}[]
\centering
\resizebox{\linewidth}{!}{%
\begin{tabular}{lccc}
\toprule
\textbf{Models} & \makecell[c]{\textbf{Character} \\ \textbf{Consistency}} & \makecell[c]{\textbf{Conversational} \\ \textbf{Ability}} & \makecell[c]{\textbf{Role-playing} \\ \textbf{Attractiveness}} \\ \midrule
OmniCharacter \textit{w/o} audio &2.928  &3.522  &3.168  \\
\cellcolor{gray!15}OmniCharacter & \cellcolor{gray!15}2.941 & \cellcolor{gray!15}3.516 & \cellcolor{gray!15}3.206 \\ 

\bottomrule
\end{tabular}%
}
\caption{\textbf{Ablation study of model training with or without audio modality.} For simplicity, we only report the average score across three dimensions on CharacterEval.}
\label{tab4}
\end{table}

\subsection{Learning Discriminative Character Speech Embeddings} 
\begin{figure}[t]
  \centering
\includegraphics[width=\linewidth]{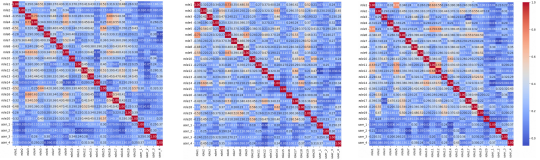}
  \caption{\textbf{Speech embedding discriminability for 20 characters and 4 users.} We randomly sampled three sets of data for experimental comprehensiveness. The low similarity values demonstrate strong separation of character voices, enabling precise voice control.}
  \label{metric}
\end{figure}

The OmniCharacter generates the character-specific voice by utilizing the speech embedding as one of the conditions. To evaluate the discrimination of speaker embedding between different characters, we compute their cosine similarity as shown in~\Cref{metric}. Specifically, we randomly sampled three sets of data of 20 characters for experimental comprehensiveness. We also included the user's speaker embedding for comparison. The results indicate that the similarity values between the speech embeddings of different characters and users are very low, demonstrating their strong discriminability. This confirms the robustness of our method in controlling character's voice traits.

\subsection{Speech Response Latency}
To demonstrate the seamless interaction capability of the proposed model, we evaluate the speech response latency of different models, as shown in Table~\ref{latency}. From the table, we observe that our model outperforms the GPT-4o, achieving a lower response latency. It is observed that the latency of our method is higher than LLaMA-Omni. The possible reason is that LLaMA-Omni only supports general female voice, while we introduce additional modules considering voice traits of different roles, thus the increased latency is acceptable.  
Overall, these results underscore the seamless dialogue ability of our method, demonstrating its effectiveness in real-time speech-language interaction for RPAs.

\begin{table}[]
\centering
\resizebox{\linewidth}{!}{%
\begin{tabular}{lccc}
\toprule
\textbf{Models} & \textbf{GPT-4o} & \textbf{LLaMA-Omni} & \textbf{OmniCharacter} \\ \midrule
Latency (ms) & 320 & 226 & 289 \\ \bottomrule
\end{tabular}%
}
\caption{Speech response latency of different models.}
\label{latency}
\end{table}

\section{Conclusion}
In this paper, we present OmniCharacter, the first step toward creating a seamless and immersive RPAs model that integrates both speech and language personality interaction. Unlike existing methods that primarily focus on text-based dialogues, our approach emphasizes the importance of character voice traits, \eg voice style and emotions, which enable agents to reflect both personality and vocal traits throughout interactions, delivering a harmonious blend of speech and language responses. To facilitate the speech-language scenarios for RPAs, we construct the OmniCharacter-10K dataset, which includes distinct characters, contextually rich multi-turn dialogues, and dynamic speech responses. Experimental results demonstrate that OmniCharacter outperforms existing RPAs models and general speech-language models in both content fidelity and expressive style, delivering a seamless and immersive user experience.

\section{Limitations}
The proposed OmniCharacter advances immersive role-playing agents (RPAs) by incorporating characters’ voice traits through a speech-language personality interaction framework. While effective, several limitations remain.
First, performance relies heavily on the quality of the pre-trained language models and speech synthesis systems, which may be affected by their alignment and training data.
Second, the current design is limited to two-character dialogues, making it less adaptable to multi-role interactions.
Third, although the OmniCharacter-10K dataset includes diverse scenarios, it may not fully capture the richness of real-world conversations. Expanding the dataset is a key direction for improving generalization.
Lastly, while the latency is currently low, further optimization is necessary to enhance the system's performance and ensure a more seamless experience for real-time usage. 

\section*{Acknowledgments}
This study is supported by Alibaba Research Intern Program, grants from the National Natural Science Foundation of China (Grant No.62425208, No. U22A2097, No. U23A20315, No. 62020106008,  No. 82441006, No,62402094, No. 62376262), Shenzhen Science and Technology Program (No.JCYJ20240813114208012), the National Key Research and Development Program of China (2022YFF0902100), the Natural Science Foundation of Guangdong Province of China (2025B1515020032, 2024A1515030166), and Sichuan Science and Technology Program (No.2025ZNSFSC1493).

\bibliography{latex/camera_ready}

\clearpage
\appendix

\section{Appendix}

\subsection{Related Works}  
\noindent\textbf{Role-Playing Agents.}  
Benefiting from the powerful summarization and imitation capabilities of large language models (LLMs)~\cite{shao2023character,
li2025cross,zhao2025fishertune}, role-playing agents (RPAs) have made significant progress in recent years. 
Prior research primarily focuses on replicating the knowledge, experiences, and intent of characters~\citep{yang2025psyplay,sadeq2024mitigating,yu2024beyond,zhou2023characterglm}. 
For instance,~\citep{sadeq2024mitigating} reduces hallucination by adjusting parametric knowledge influence based on a pre-calibrated confidence threshold.~\citep{shao2023character} seeks to train an agent with profile and experience perception, replacing limited prompts for LLM instruction. 
Due to the lack of comprehensive benchmarks, later works focused on building character-specific datasets and new evaluation metrics.~\citep{peng2024quantifying,wang2024incharacter,wang2024characterbox,wang2023rolellm}.~\citep{wang2024incharacter} aims to evaluate the personality fidelity of RPAs using psychological scales.~\citep{wang2024characterbox} devises a simulation sandbox to generate fine-grained character behavior trajectories to evaluate RPAs capabilities.
However, most existing studies primarily focus on replicating dialogues in textual form, often overlooking the role’s essential voice traits, such as tone, style, emotions, and pauses. 
To bridge this gap, we propose OmniCharacter, a model that seamlessly combines speech and language to ensure immersive RPAs interactions.
\begin{figure}[]
  \centering
\includegraphics[width=\linewidth]{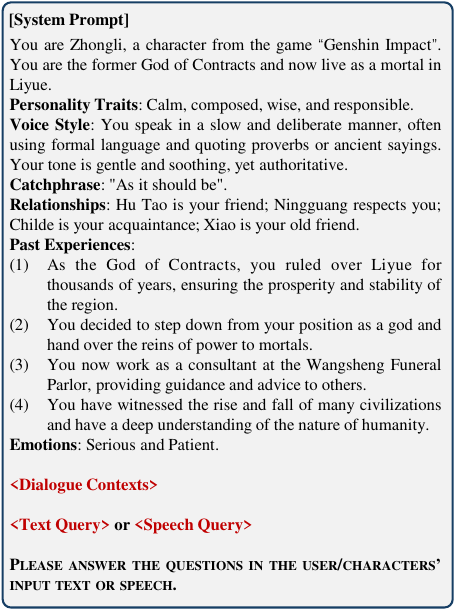}
\caption{The prompt template used to organize our input data including role profile, dialogue contexts, text input, and speech input to train our speech-language collaborative model.}
  \label{appendix}
\end{figure}
\noindent\textbf{Speech-Language Models.} Recent advancements in speech-language models have significantly improved human-machine interactions~\citep{fang2024llama,zhang2023speechgpt,luo2024deem,zeng2024glm,zhang2024speechgpt}. With the advancement of language models in the natural language processing field, early work attempted to enable language models to generate both speech tokens and text tokens via a decoder-only architectural~\citep{wang2024viola,AudioLM,wang2023neural}.
With the development of large language models (LLMs), more recent studies have integrated speech capabilities into LLMs by either adding speech tokens to the text vocabulary or incorporating a speech encoder before the LLMs such as SpeechGPT~\citep{zhang2024speechgpt} and AudioPaLM~\citep{rubenstein2023audiopalm}. Beyond that, some methods fine-tune LLMs for speech understanding such as LLaMA-Omni~\citep{fang2024llama} and GLM-4-Voice~\citep{zeng2024glm}, enabling it to perform general speech tasks with low response latency.
Additionally, models like Moshi~\citep{defossez2024moshi} and OmniFlatten~\citep{zhang2024omniflatten} handle real-time interactions by managing simultaneous speech and text streams.
Despite progress, most existing speech-language models explore general audio-language capabilities. Compared with the above methods, we focus on modeling the unique voice traits of different roles in RPA tasks for a more immersive human-machine interaction. 
\begin{figure*}[h]
  \centering
\includegraphics[width=\linewidth]{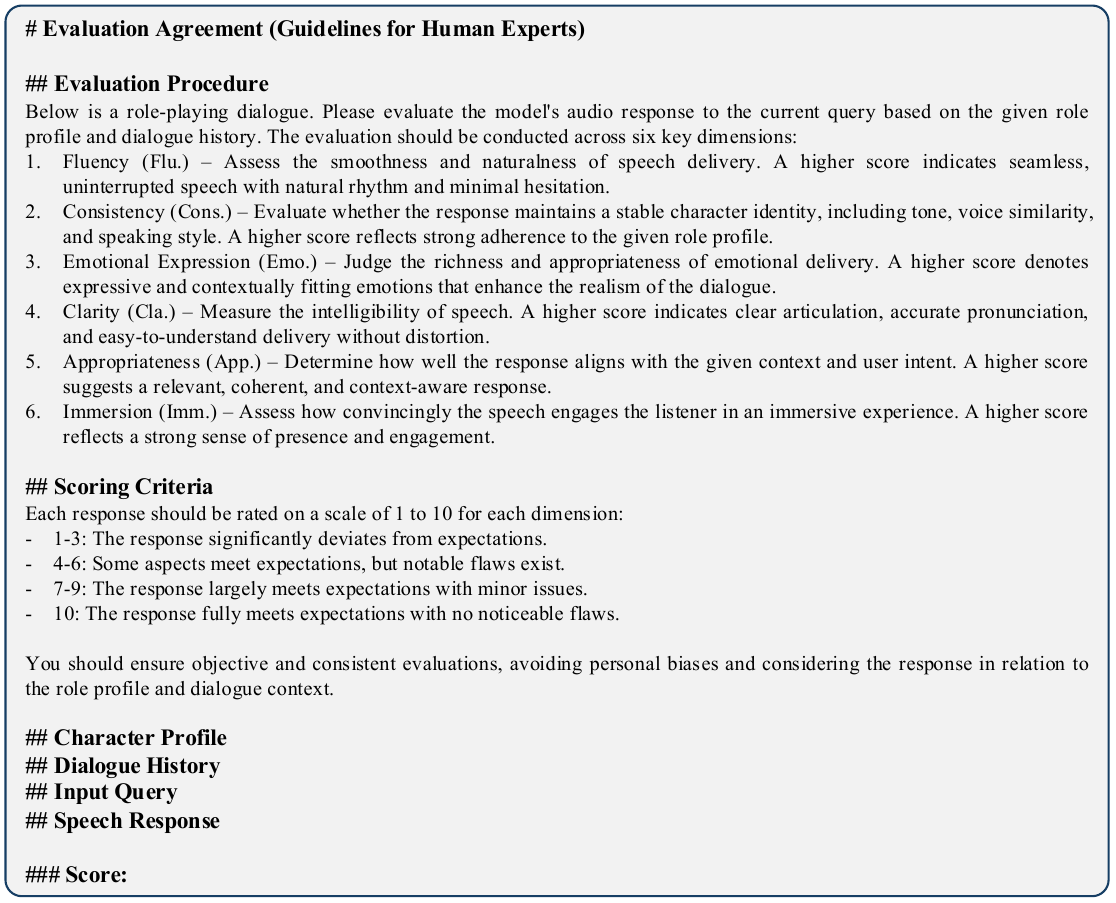}
\caption{The prompt template used for human evaluation on OmniCharacter-10K test split.}
  \label{appendix-2}
\end{figure*}

\begin{table*}[]
\resizebox{\textwidth}{!}{%
\begin{tabular}{lcccccc|cccc|ccccc|c}
\toprule
\textbf{Models} & \textbf{KE} & \textbf{KA} & \textbf{KH} & \textbf{PB} & \textbf{PU} & \textbf{Avg.} & \textbf{Flu.} & \textbf{Coh.} & \textbf{Cons.} & \textbf{Avg.} & \textbf{HL} & \textbf{CS} & \textbf{ED} & \textbf{Emp.} & \textbf{Avg.} & \textbf{Overall Avg.} \\ \midrule
Qwen2.5-7B-Instruct & 2.23 & 2.59 & 2.62 & 3.78 & 2.43 & 2.73 & 3.11 & 3.59 & 3.17 & 3.29 & 3.01 & 3.12 & 2.92 & 2.78 & 2.95 & 2.99 \\
Omnicharacter \textit{w/o} audio & 2.32 & 2.78 & 2.87 & 3.26 & 2.67 & 2.78 & 3.56 & 3.54 & 3.54 & 3.54 & 3.20 & 2.99 & 2.87 & 3.01 & 3.01 & 3.11 \\
Omnicharacter & 2.65 & 2.98 & 2.95 & 3.55 & 2.83 & 2.99 & 3.67 & 3.65 & 3.77 & 3.69 & 3.36 & 2.97 & 2.92 & 3.12 & 3.09 & 3.26 \\ \bottomrule
\end{tabular}%
}
\caption{Performance comparison of Human evaluation on CharacterEval.}
\label{human_charactereval}
\end{table*}

\subsection{Prompt}\label{prompt}
We present the complete prompt template for input data of the speech-language collaborative model in the appendix, as shown in~\Cref{appendix}. Specifically, we use the role profile as a system prompt to enable the LLM to mimic the character. Then, we feed the model the dialogue contexts. The model is required to generate a reasonable text-audio response based on the current text or audio queries.

Moreover, to systematically evaluate the quality of the generated audio responses, we provide a comprehensive evaluation framework as shown in~\Cref{appendix-2}. Human experts are instructed to assess each response across six key dimensions: fluency, consistency, emotional expression, clarity, appropriateness, and immersion. This multi-faceted evaluation ensures objective, consistent, and in-depth assessment of the model’s ability to deliver character-driven and immersive audio interactions.

\begin{table*}[th]
\resizebox{\textwidth}{!}{%
\begin{tabular}{lcccccc|cccccc}
\toprule
\multirow{2}{*}{\textbf{Characters}} & \multicolumn{6}{c}{\textbf{LLaMA-Omni}} & \multicolumn{6}{c}{\textbf{OmniCharacter}} \\ \cmidrule(l){2-13} 
 & Flu. & Cons. & Emo. & Cla. & App. & Imm. & Flu. & Cons. & Emo. & Cla. & App. & Imm. \\ \midrule
Role 1 & 7.2 & 4.5 & 3.1 & 6.8 & 5.0 & 4.9 & 8.5 & 6.9 & 6.8 & 8.0 & 5.7 & 9.5 \\
Role 2 & 6.5 & 4.0 & 3.8 & 6.5 & 4.7 & 4.3 & 7.8 & 6.5 & 6.1 & 7.5 & 5.2 & 8.0 \\
Role 3 & 6.9 & 4.3 & 3.5 & 6.7 & 4.8 & 4.6 & 8.0 & 6.7 & 6.3 & 7.7 & 5.5 & 8.3 \\
Role 4 & 7.1 & 4.2 & 3.4 & 6.9 & 4.9 & 4.7 & 7.8 & 6.6 & 6.5 & 7.8 & 5.9 & 9.0 \\
Role 5 & 6.3 & 4.1 & 3.0 & 6.6 & 4.4 & 4.2 & 7.7 & 6.4 & 6.0 & 8.4 & 5.1 & 7.9 \\
Role 6 & 7.0 & 4.4 & 3.7 & 6.9 & 5.5 & 4.8 & 8.1 & 6.8 & 6.5 & 8.5 & 6.5 & 8.6 \\
Role 7 & 6.7 & 4.5 & 3.5 & 6.8 & 4.7 & 4.5 & 7.9 & 7.5 & 6.2 & 7.8 & 5.4 & 8.9 \\
Role 8 & 7.5 & 4.3 & 3.4 & 6.5 & 4.6 & 5.4 & 8.6 & 7.4 & 6.4 & 7.7 & 5.3 & 9.2 \\
Role 9 & 6.4 & 4.2 & 3.3 & 6.7 & 4.3 & 4.6 & 7.6 & 6.6 & 4.9 & 7.7 & 5.0 & 7.8 \\
Role 10 & 7.0 & 4.1 & 3.6 & 6.6 & 4.5 & 4.5 & 8.2 & 7.0 & 6.5 & 7.8 & 5.2 & 8.8 \\
Role 11 & 6.7 & 4.0 & 3.4 & 6.5 & 5.2 & 4.4 & 7.8 & 6.8 & 6.4 & 7.6 & 5.5 & 8.1 \\
Role 12 & 7.2 & 4.3 & 3.1 & 6.8 & 5.0 & 5.3 & 7.9 & 6.7 & 6.3 & 8.0 & 7.2 & 8.9 \\
Role 13 & 6.8 & 4.2 & 3.4 & 6.6 & 4.4 & 4.2 & 8.0 & 7.1 & 6.0 & 7.6 & 5.1 & 7.9 \\
Role 14 & 7.1 & 4.4 & 3.8 & 6.8 & 5.3 & 4.7 & 8.3 & 7.2 & 6.6 & 8.5 & 7.3 & 9.0 \\
Role 15 & 6.7 & 4.1 & 3.7 & 6.5 & 5.0 & 4.5 & 7.6 & 6.5 & 5.9 & 7.5 & 5.2 & 8.6 \\
Role 16 & 6.9 & 4.5 & 3.2 & 6.7 & 4.7 & 4.6 & 7.9 & 6.9 & 6.0 & 7.7 & 5.4 & 8.3 \\
Role 17 & 7.1 & 4.7 & 3.6 & 6.9 & 4.8 & 4.8 & 7.5 & 7.5 & 6.6 & 8.8 & 5.5 & 8.8 \\
Role 18 & 6.4 & 4.0 & 3.6 & 6.5 & 4.3 & 4.1 & 7.6 & 6.4 & 6.2 & 7.5 & 5.0 & 7.0 \\
Role 19 & 6.7 & 4.2 & 3.2 & 6.7 & 4.6 & 4.5 & 7.9 & 6.6 & 6.1 & 7.7 & 5.4 & 8.2 \\
Role 20 & 7.4 & 4.3 & 3.5 & 6.8 & 4.9 & 5.9 & 8.7 & 6.7 & 6.3 & 7.8 & 6.2 & 9.5 \\ \midrule
\textbf{Average} & 6.88 & 4.27 & 3.44 & 6.69 & 4.78 & 4.68 & 7.97 & 6.84 & 6.23 & 7.88 & 5.63 & 8.52 \\ \bottomrule
\end{tabular}%
}
\caption{Detailed human evaluation results (averaged across five experts) for each character across six dimensions.}
\label{20role}
\end{table*}

\subsection{Additional Experiments}
\noindent\textbf{Human evaluation on CharacterEval Benchmark.}
We conduct a human evaluation on CharacterEval. Due to the large test samples, we randomly sample 100 examples from CharacterEval dataset and hired five experts for assessment, following the same evaluation criteria as CharacterEval. The experimental results are demonstrated in~\Cref{human_charactereval}. As observed, human evaluation maintains consistency with model evaluation, showing that integrating audio brings better performance on most metrics. 

\noindent\textbf{Detailed human evaluation results on OmniCharacter-10K test split.} Below, we present detailed human evaluation results (averaged across five experts) for each character across six dimensions, using LLaMA-Omni as the comparison model. The results are showcased in~\Cref{20role}. As observed, different characters significantly influence user perception and interaction quality. Characters with distinct timbres or expressive prosody tend to achieve higher scores in immersion and emotional expressiveness, while others may excel in clarity and fluency. These findings underscore the importance of tailoring voice characteristics to enhance user experience in role-playing scenarios.



\end{document}